\pdfoutput=1

\documentclass[11pt]{article}

\usepackage[preprint]{acl}
\usepackage{amssymb}
\usepackage{xcolor}
\usepackage{times}
\usepackage{latexsym}
\usepackage{algorithm}
\usepackage{algpseudocode}
\usepackage{multirow}
\usepackage{tabularx}
\usepackage{afterpage}
\usepackage{amsmath}
\usepackage{booktabs}

\usepackage[most]{tcolorbox}

\usepackage[T1]{fontenc}

\usepackage[utf8]{inputenc}

\usepackage{microtype}

\usepackage{inconsolata}

\usepackage{graphicx}
\usepackage{makecell}

\DeclareMathOperator*{\argmin}{arg\,min}

\title{Positional task conditioning for scalable defect detection across product families in large product catalogs}

\author{
 \textbf{Soham Satyadharma\textsuperscript{1}},
 \textbf{Gabriel Roccabruna\textsuperscript{1}},
 \textbf{Suleiman A. Khan\textsuperscript{1}}
\\
 \textsuperscript{1} Amazon Catalog AI
\\
 \small{
 {ssatyadh@amazon.com, groccabr@amazon.es, suleimkh@amazon.com}
 }
}

\begin{document}
\maketitle
\begin{abstract}

Product families in large product catalogs suffer from inconsistencies such as duplicates and unit mismatches that degrade customer experience. Detecting these requires reasoning over multiple error types across lengthy product listings, where LLM classification quality degrades due to long-context limitations. We address this by decomposing detection into focused sub-tasks that reduce context and isolate error types, improving F1 from 52\% to 87\%. For scalable deployment, we introduce Positional Task Conditioning (PTC), which distills this capability into a single smaller model by reinforcing task identity at structural prompt boundaries. PTC outperforms rationale-based distillation across five models and two architecture families, achieving within 1.79\% F1 of the frontier at upto 98\% lower cost. Our system is deployed across multiple countries processing 10+ million product families.

\end{abstract}

\section{Introduction}
\label{sec:introduction}
Large product catalogs organize products into logical groupings that differ by attributes like size, color, or style. They suffer inconsistencies from heterogeneous seller inputs and automated ingestion manifesting as duplicate variations, unit conflicts, attribute overstuffing, and theme defects \cite{schmidts2020catalog}. These inconsistencies directly degrade customer experience: duplicates (e.g., \textit{XL} vs. \textit{Extra Large}) increase cognitive load, unit conflicts (e.g., \textit{Medium} vs. \textit{30 inches}) force manual conversions, and overstuffed attributes conflate orthogonal concepts. Poor product information has endemically increased return rates and support costs while reducing conversions \cite{xu2015will}, making automated detection critical for large scale operational efficiency.

Prior work addresses single-product attribute correctness \cite{wang2020automatic}, entity matching \cite{li2020deep}, or attribute generation \cite{nikolakopoulos2023sage}, leaving intra-family inconsistency detection underexplored. Classical rule-based approaches lack generalization \cite{ilyas2019data}, and manual auditing is intractable at scale, necessitating automated solutions. 

A natural approach is providing complete family data and guidelines to a Large Language Model (LLM) to identify all errors simultaneously. However, families with numerous products create lengthy contexts where LLMs struggle to utilize the correct information \cite{liu2024lost}. Multi-label classification requiring simultaneous reasoning over orthogonal error types challenges state-of-the-art models \cite{ kou2025rethinking, ortego2025largelanguagemodelsmeet, tabatabaei2025can}. Inspired by work showing task decomposition improves LLM reasoning \cite{khot2022decomposed, zhou2022least}, we adopt a divide-and-conquer strategy: we structure detection as nested attribute-level and task level classification with dedicated prompts, enabling focused reasoning and reduced context, to generate final predictions of egregious, non-egregious, or no error. 

\begin{figure}
    \centering
    \includegraphics[width=0.98\columnwidth]{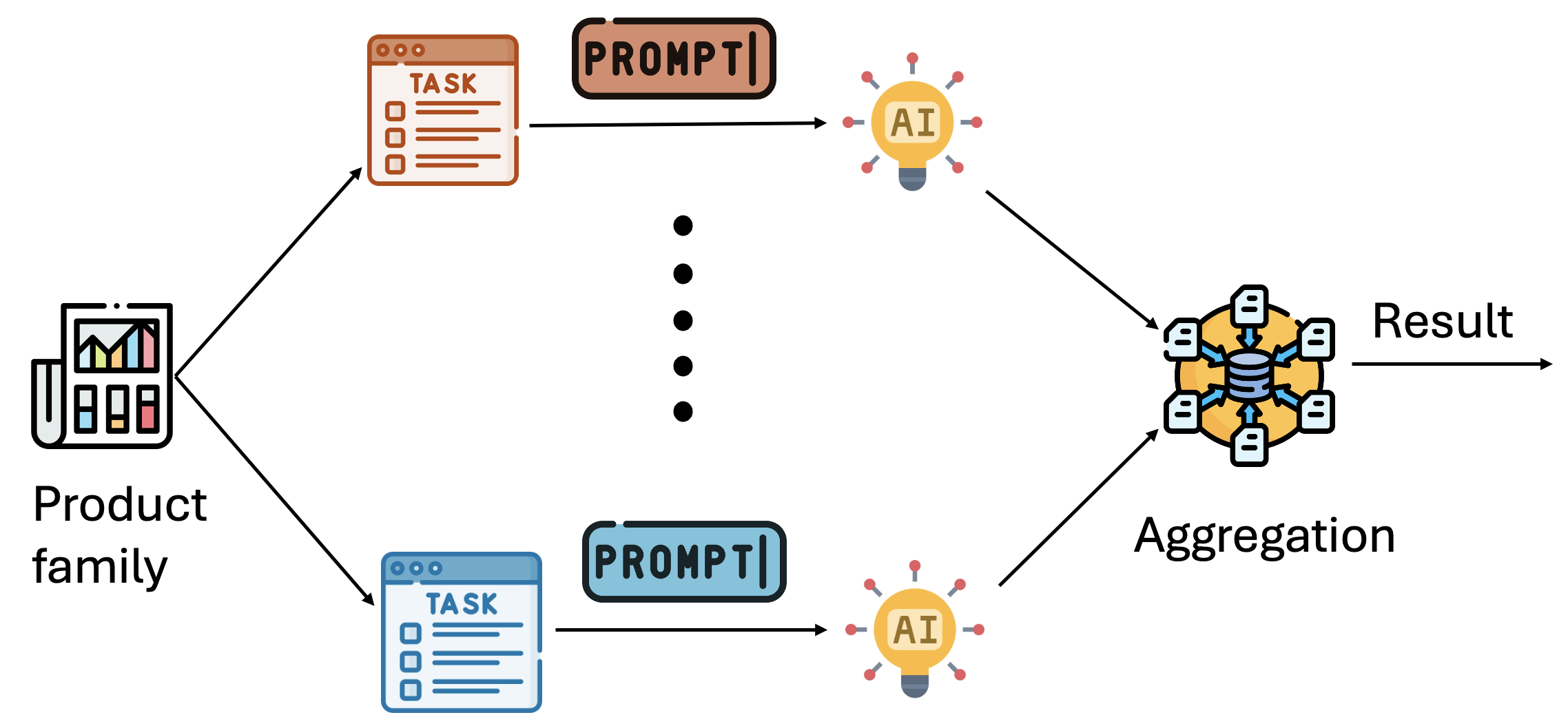}
     \caption{The figure depicts our divide and conquer approach. We take a product family as input and decompose each of its variating attributes (\S \ref{sec:problem_formulation}) into multiple classification tasks. Each task has its own prompts; the results are aggregated for the family.}
    \label{fig:end_to_end}
\end{figure}


While our decomposition achieves production-grade accuracy, frontier LLMs at catalog scale incur prohibitive costs. Knowledge distillation enables transferring capabilities to efficient student models \cite{hinton2015distilling, gou2021knowledge}, but distilling separate models for each task introduces deployment complexity. Rationale-based distillation (RBD) \cite{hsieh2023distilling} has demonstrated multi-task capabilities by training on teacher reasoning chains, but a gap between student and teacher performance persists. We propose Positional Task Conditioning (PTC), a task conditioning strategy for distillation, that bridges this gap by conditioning on task identifiers at structural prompt boundaries, leveraging context enrichment to improve generalization \cite{caruana1997multitask}. 

Our contributions are: (1) a categorization of variation family inconsistencies spanning four tasks with different severity levels; (2) a divide-and-conquer framework decomposing multi-label detection into three-class sub-tasks, improving precision by 54\% and recall by 13\% over monolithic prompting; (3) Positional Task Conditioning (PTC), a multi-task distillation methodology enabling a single model to handle all sub-tasks, outperforming rationale-based distillation while approaching frontier performance at upto 98\% reduced cost; and (4) empirical validation on both human-annotated and synthetic data demonstrating PTC generalizes across models, yielding consistent improvements across five LLMs from two architecture families.

\section{Related work}
\label{sec:related_work}
\paragraph{Multi-label classification and LLMs.} The paradigm for multi-label text classification has shifted from discriminative methods \cite{devlin2019bert, liu2019roberta} to generative LLMs, which often violate output space constraints \cite{niraula2024multi} and fail to capture conditional dependencies in flat generation settings \cite{ma2025large, zhou2024quest}. We address this by decomposing the monolithic output space into four modular three-class classification sub-tasks.

\paragraph{Multi-task learning.} Multi-task learning (MTL) improves generalization by leveraging shared structure across related tasks \cite{ruder2017overview}, traditionally via a shared encoder with task-specific heads \cite{liu2019multi}. PTC departs from this paradigm: instead of task-specific heads, it conditions a single LLM across all sub-tasks by inserting task identifiers at structural prompt boundaries, enabling one student model to discriminate between structurally similar prompts without any architectural changes.

\paragraph{Distillation in LLMs.} While frontier LLMs exhibit emergent reasoning via CoT prompting \cite{wei2022chain, kojima2022large}, their deployment is constrained by prohibitive latency and cost. Knowledge Distillation \cite{hinton2015distilling} addresses this by transferring capabilities to a student model, with recent approaches training students on intermediate reasoning traces \cite{wang2023scott, magister2023teaching, hsieh2023distilling, mukherjee2023orca, trabelsi2025matters}. 

We extend this paradigm by introducing positional task conditioning that inserts task identifiers at multiple structural boundaries within the prompt. Unlike T5-style single-prefix conditioning \cite{raffel2020exploring}, which prepends a task token only at sequence start, PTC reinforces task identity at semantically meaningful positions. Related work on prompt optimization across foundation models 
\cite{opsahl2024optimizing, gao2025prompt, venkataraman2026framework} similarly emphasizes that prompt structure materially affects model behavior, though we focus on positional task conditioning for distillation rather than prompt translation. Our ablations over seven configurations show that multi-position placement significantly outperforms the single-prefix baseline.

\paragraph{LLMs in product catalogs.} LLMs have been deployed across large product catalogs for recommendation \cite{maragheh2023llm}, search \cite{rokon2026enhancementecommercesponsoredsearch}, product discovery \cite{wang2024leveraging}, knowledge graph completion \cite{chen2023knowledge}, product matching \cite{herrero2024learning}, and categorization \cite{cheng2024commerce, kathiriya2023optimizing}. Recent efforts extend to attribute extraction \cite{baumann2024using, zhang2025leveraging, mansoori2026improving} and single-product quality assessment \cite{satyadharma2025auto}, but to our knowledge no prior work addresses product family quality detection at large scale. A related challenge is obtaining high-quality annotations at volume; we adopt the approach of generating synthetic labels by injecting errors into correct inputs \cite{negri2025attributeawarecontrolledproductgeneration}, enabling scalable evaluation data generation without manual labeling.

\section{Method}
\label{sec:method}

\subsection{Problem formulation}
\label{sec:problem_formulation}

We formalize the task of detecting inconsistencies in product families as a multi-label classification problem. Let $\mathcal{F}$ denote the space of all families in a product catalog. A family $F \in \mathcal{F}$ consists of a set of $n$ products $F = \{P_1, P_2, \ldots, P_n\}$, where each product $P_i$ is represented as a set of attribute-value pairs: $P_i = \{(a_1, v_1), (a_2, v_2), \ldots, (a_m, v_m)\}$. Here, $a_j \in \mathcal{A}$ denotes an attribute from the catalog's attribute schema (e.g., \texttt{size}, \texttt{color}, \texttt{style}), and $v_j$ represents the corresponding string value for attribute $a_j$. We define the \textit{variating attributes} of a family as the set of attributes along which family members differ: $V(F) = \{a_j \in \mathcal{A} : |\{v_j^{(i)} : P_i \in F\}| > 1\}$. For instance, a family of t-shirts varying in size and color would have $V(F) = \{\texttt{size}, \texttt{color}\}$.

To judge the quality of a product family, we define a set of error types $\mathcal{E}$, where each error type $e_i$ is associated with natural language instructions $I_i$ specifying detection criteria, including definitions, edge cases, and few-shot examples. The complete input for a family $F$ is $X = (F, V(F), \mathcal{I})$ where $\mathcal{I} = \{I_1, I_2, \ldots, I_{|\mathcal{E}|}\}$. The detection task requires learning a function:

\begin{equation}
    f: \mathcal{X} \rightarrow 2^{\mathcal{E} \times \mathcal{A}}
\label{eqn:multi_label}
\end{equation}

that maps input to a subset of (error type, attribute) pairs. The input comprising instructions for all error types and multiple product records, leads to long context, challenging LLM accuracy.

\subsection{Divide and conquer (D\&C) approach}
\label{sec:divide_and_conquer}
Based on semantic equivalence of error types in $\mathcal{E}$, we decompose $f$ (Eqn.~\ref{eqn:multi_label}) into a set of $|\mathcal{T}|$ focused sub tasks, each with its own labels. Let the combined label set be $\mathcal{Y}$. Each task groups semantically related error types through its labels: we define a labeling map $\psi: \mathcal{T} \times \mathcal{Y} \rightarrow \mathcal{E} \cup \{\varnothing\}$, where $\psi(t_k, y)$ is the error type when task $t_k$ emits label $y$, and $\varnothing$ when $y$ denotes \textit{no error}. 
This enables us to reformulate detection as $|\mathcal{T}|$ independent classification problems rather than a single $|\mathcal{T}| \times |\mathcal{Y}|$ multi-label problem. For each task $t_k \in \mathcal{T}$ and variating attribute $a \in V(F)$, we define a sub-task classifier $g_{k}: (F, a, I'_k) \rightarrow \mathcal{Y}$, where $I'_k$ is the instruction set for $t_k$. Each sub-task operates on reduced context: only the attribute values $\{v_a^{(i)} : P_i \in F\}$ and instructions for a single task.

Each sub-task $g_k$ is instantiated via a tailored prompt $x^{(k)}$ comprising four components:
(1) \emph{Introduction:} Role definition establishing the model as a catalog quality expert. 
(2) \emph{Instructions $I'_k$:} Detailed criteria for task $t_k$ and attribute $a$, including definitions, boundary cases, and few-shot examples.
(3) \emph{Family input:} The set of values $\{v_a^{(i)} : P_i \in F\}$, for attribute $a$.
(4) \emph{Output schema:} Structured output response

We assemble the family level prediction as $f(X) = \bigcup_{k=1}^{|\mathcal{T}|} \bigcup_{a \in V(F)} \phi(g_k(F, a, I'_k), t_k, a)$, where $\phi: \mathcal{Y} \times \mathcal{T} \times \mathcal{A} \rightarrow 2^{\mathcal{E} \times \mathcal{A}}$ maps a sub-task output to the corresponding (error type, attribute) pair via $\psi$  where $\phi(y, t_k, a) = \{(\psi(t_k, y),\, a)\}$ for defect labels and $\emptyset$ otherwise. 
This decomposition requires at most $|\mathcal{T}| \times |V(F)|$ LLM invocations per family, but with reduced context and a simpler decision boundary per call.

\subsection{Positional task conditioning}
\label{sec:positional_task_conditioning}
While D\&C achieves strong performance with frontier LLMs, scalable deployment incurs prohibitive costs. We address this through distillation into a single small model. We first formalize a Rationale-Based Distillation baseline (RBD), then present our Positional Task Conditioning (PTC) approach, which extends it with explicit task conditioning at structural prompt boundaries.

Let $M_\theta$ denote a student LLM parameterized by $\theta$. For each task $t_k \in \mathcal{T}$, we construct training dataset $\mathcal{D}_k = \{(x_i^{(k)}, y_i^{(k)})\}_{i=1}^{N_k}$ from teacher model outputs on sub-task $g_k$. Each input $x_i^{(k)} = (F_i, a_i, I'_k)$ comprises a family, attribute, and task specific instructions, with label $y_i^{(k)} \in \mathcal{Y}$. Each input has an inherent structure $x_i^{(k)} = [s_{\text{intro}}, s_{\text{inst}}, s_{\text{input}}, s_{\text{output}}]$ corresponding to the prompt components defined in Sec.~\ref{sec:divide_and_conquer}.

Following research on distilling reasoning traces from larger models \cite{ho2023large, fu2023specializing}, we supervise the student with teacher-generated reasoning chains. For each sub-task $g_k$, the teacher model $M'$ produces a reasoning chain and label $(r_i^{(k)}, y_i^{(k)})$ and we train on these sampled outputs. RBD minimizes the negative log-likelihood over all sub-tasks, $\theta_{\text{rbd}} = \argmin_{\theta} \sum_{k=1}^{|\mathcal{T}|} \sum_{i=1}^{N_k} -\log P_\theta(r_i^{(k)}, y_i^{(k)} \mid x_i^{(k)})$. However, RBD lacks explicit task awareness: the model must implicitly infer the task from the instructions, leading to confusion when prompts share a similar structure across tasks.

\begin{figure} 
    \centering 
    \includegraphics[width=\columnwidth]{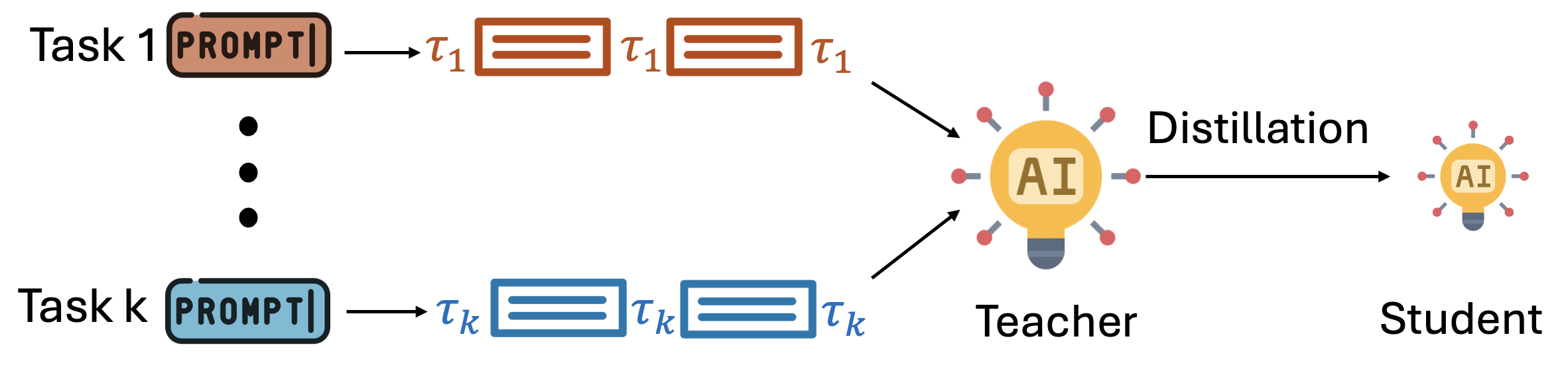} 
    \caption{Positional Task Conditioning. Task tokens ($\tau_1, \ldots, \tau_k$) are inserted at structural boundaries within each prompt: before the introduction, before the input data, and at the end. The teacher LLM data is generated using these task tokens and used to distill the student LLM.} 
    \label{fig:ptc} 
\end{figure}

PTC (Fig \ref{fig:ptc}) introduces explicit task conditioning by associating each task $t_k$ with dedicated tokens $\tau_k$ and inserting $\tau_k$ at structurally meaningful boundary positions: before the introduction, before the input data, and after the output schema. The positionally-conditioned input is: 
\begin{equation}
    \tilde{x}_i^{(k)} = \tau_k \oplus s_{\text{intro}} \oplus s_{\text{inst}} \oplus \tau_k \oplus s_{\text{input}} \oplus s_{\text{output}} \oplus \tau_k 
\end{equation}

 where $\oplus$ denotes concatenation. PTC optimizes 
 \begin{equation}
     \theta_{\text{ptc}} = \argmin_{\theta} \sum_{k=1}^{|\mathcal{T}|} \sum_{i=1}^{N_k} -\log P_\theta(r_i^{(k)}, y_i^{(k)} | \tilde{x}_i^{(k)})
 \end{equation}
 .

We hypothesize that each placement plays a complementary role: the initial $\tau_k$ primes the task before the prompt is read, the mid-prompt $\tau_k$ re-anchors task identity at the transition from instructions to data, and the final $\tau_k$ marks the transition into the generation phase. We test this hypothesis empirically in \S \ref{sec:experimental_setup} across 7 positional configurations.

\subsection{Synthetic data pipeline}
\label{sec:synthetic_data_pipeline}

To generate evaluation data at scale, we design a synthetic pipeline that produces families with controlled error injections. 
Given a family $F$ (\S ~\ref{sec:problem_formulation}), each product $P_i$ has an associated textual description $d_i$. For each attribute $a \in V(F)$, we first generate a clean list of values by prompting an LLM with a sanitized description (brand-anonymized, attribute-references removed) and attribute-specific formatting instructions: $v_{\text{gen}}$. 
We then inject errors via type-specific functions $\text{Inj}_k$ for each task $t_k \in \mathcal{T}$, producing $v_{\text{defect}} = \text{Inj}_k(v_{\text{gen}})$. Injection strategies are prompt-based (e.g., generating semantic duplicates) or deterministic (e.g., mixing unit systems). To avoid cross-interference when injecting multiple error types into the same family, we create an injection planner $ac_k = (\text{conditions}, \text{effects}, \text{Inj}_k)$ to find a valid injection order respecting inter-error dependencies.

\section{Experimental setup}
\label{sec:experimental_setup}

We address four research questions: \textbf{RQ1:} Does divide-and-conquer outperform monolithic prompting? \textbf{RQ2:} Does PTC outperform prompting and RBD? \textbf{RQ3:} Does PTC generalize across models? \textbf{RQ4:} How does task token placement affect performance?

\paragraph{Tasks.}
\begin{figure} 
    \centering 
    \includegraphics[width=\columnwidth]{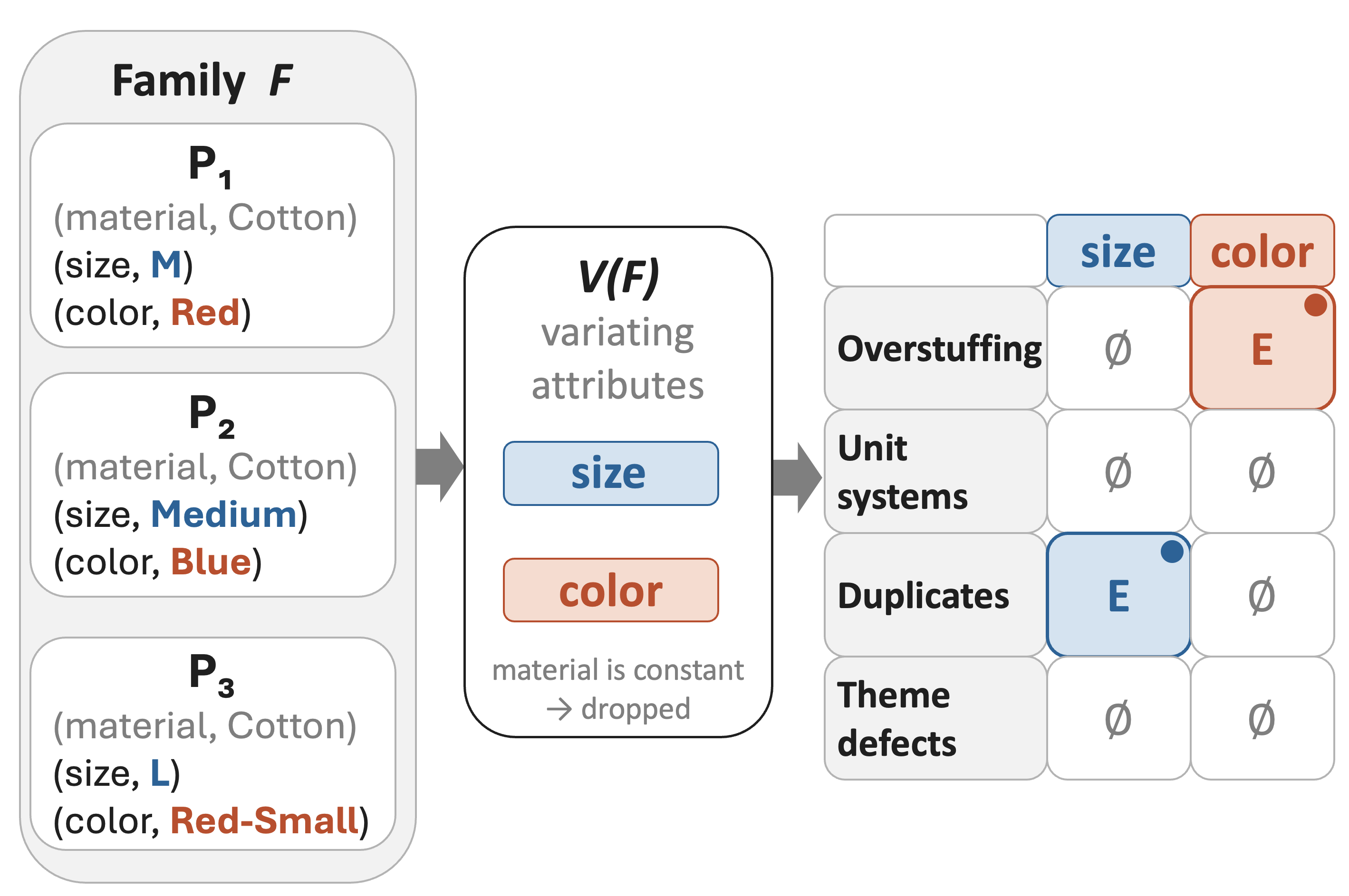} 
    \caption{Divide-and-conquer on a t-shirt family $F$: we extract the variating attributes $V(F)$ (dropping the constant \texttt{material}) and classify each (task, attribute) pair independently as no error ($\phi$), non-egregious (NE), or egregious (E). Two egregious defects are detected in this example: \textit{Overstuffing} on \texttt{color} and \textit{Duplicates} on \texttt{size}.} 
    \label{fig:variation_quality} 
\end{figure}
To comprehensively model the quality of a product family, we decompose eight error types $\mathcal{E}$ into four tasks, where each task is a pair of semantically related egregious (E) and non-egregious (NE) error variant. Hence, our D\&C approach becomes four independent classification problems, each with three labels: \textit{egregious error}, \textit{non-egregious error}, or \textit{no error}. Our tasks are: \textit{overstuffing} ($t_1$): extraneous concepts injected into an attribute field (E: inconsistently, NE: consistently); \textit{unit systems} ($t_2$): incompatible measurement types (E: mixed systems, NE: mixed units within one system); \textit{duplicates} ($t_3$): semantically identical variations such as \textit{XL} vs.\ \textit{Extra Large} (E) or formatting inconsistencies (NE); and \textit{theme} ($t_4$): attribute values belonging to a different attribute entirely (E: inconsistently, NE: consistently). Egregious errors directly impair customer decisions; NE errors are quality concerns that do not mislead customer.  Each task token $\tau_k$ is a short natural-language marker inserted verbatim into the prompt at the positions defined in \S\ref{sec:positional_task_conditioning}; e.g., the duplicates task uses $\texttt{<TASK=DUPLICATES>}$. These markers are present in the student's training inputs, so the student learns to condition on them.

\paragraph{Baselines and methods.}
For RQ1, we compare monolithic prompting (all eight error types simultaneously) against D\&C (four separate three-class sub-tasks) using Claude Sonnet 4.5.

For RQ2, we compare three configurations: (1) a non-finetuned baseline on student models using D\&C prompts without parameter updates; (2) RBD, which finetunes on reasoning chains without explicit task conditioning; and (3) PTC, which inserts task identifiers at structural prompt boundaries.

For RQ3, we evaluate across five models spanning two architecture families: Qwen3-8B, Qwen3-14B, Qwen3-32B \cite{yang2025qwen3}, Mistral-Nemo 12B \cite{mistralnemo2024}, and Mistral-Small 24B \cite{mistralsmall24b}. RBD and PTC share identical training data and hyperparameters; for the positional ablation, only task token placement varies.

For RQ4, we perform ablations on task token placement on Qwen3-8B, exhaustively evaluating all 7 non-empty subsets of three structurally meaningful positions (start of prompt, before input, end of prompt) defined in \S \ref{sec:positional_task_conditioning} on the human labeled evaluation set detailed below.

\paragraph{Metrics and datasets.}

Our primary metric is detection of error (either E or NE) evaluated via precision, recall, and F1. We additionally analyze per task trends in the results and report full per task  metrics in the appendix. 

For training, we compile 73,723 family-attribute pairs from 45,000 product families sampled across four English-language countries (approx 14\% defect rate). Training labels are generated via distillation from Claude Sonnet 4.5 using divide-and-conquer prompting (Sec.~\ref{sec:divide_and_conquer}). We run the teacher thrice with temperature 1 and retain only unanimous labels (98.13\% of samples), discarding ambiguous cases. We use an 85/15 train/validation split stratified by error type.

For evaluation, our primary test set is a human labeled dataset of 2,296 family-attribute pairs (390 with any error; 257 overstuffing, 57 unit systems, 50 theme, 106 duplicates) with labels annotated by SMEs, verified through a 10\% blind audit achieving $\geq$98\% accuracy. \footnote{To the best of our knowledge, there are no publicly available datasets for these tasks.} Since high-quality human annotation is costly and certain error types are underrepresented in real product families, we additionally construct a synthetic test set of 11,130 family-attribute pairs (4,562 with any error; 3,181 overstuffing, 825 unit systems, 1,372 theme, 1,158 duplicates) via the pipeline in Sec.~\ref{sec:synthetic_data_pipeline}, enabling controlled ablation studies with balanced task representation. 

\paragraph{Implementation details.} We finetune using Axolotl \cite{axolotl} with QLoRA (rank 64, alpha 128, 4-bit NF4 quantization, LoRA dropout 0.05) \cite{dettmers2023qlora}, training for 3 epochs with early stopping on held-out data. All experiments run on 8x NVIDIA H100 GPUs (112-384 GPU hours per run depending on model size). The teacher model is Claude Sonnet 4.5 with D\&C prompting (\S~\ref{sec:divide_and_conquer}). We use fused AdamW (learning rate 2e-4, cosine schedule, 100 warmup steps), micro batch size 2 with gradient accumulation over 16 steps (effective batch size 32), and sequence length 8192 tokens. We use the same training data and hyperparameters for RBD and TCF.  Inference is performed using vLLM \cite{kwon2023efficient}.

\section{Results}
\label{sec:results}
\begin{table}[t]
\centering
\scriptsize
\setlength{\tabcolsep}{1pt}
\begin{tabular}{@{}cccccc@{}}
\toprule
\textbf{Model} & \textbf{Method} & \textbf{Precision} & \textbf{Recall} & \textbf{F1 score} \\
\toprule
\multirow{2}{*}{\makecell{Claude 4.5\\Sonnet}} & Monolithic & 0.4338 {\tiny($\pm$0.0498)} & 0.6535 {\tiny($\pm$0.0391)} & 0.5214 {\tiny($\pm$0.0489)} \\
& D\&C & \textbf{0.9722} {\tiny($\pm$0.0175)} & \textbf{0.7974} {\tiny($\pm$0.0402)} & \textbf{0.8762} {\tiny($\pm$0.0264)} \\
\bottomrule
\end{tabular}
\caption{Comparison of monolithic and divide-and-conquer prompting strategies on Claude 4.5 Sonnet. Decomposing the multi-label classification problem into focused sub-tasks improves F1 from 52.14\% to 87.62\%.}
\label{tab:divide_and_conquer_results}
\end{table}

\paragraph{Divide-and-conquer effectiveness.}
Table~\ref{tab:divide_and_conquer_results} compares monolithic and D\&C prompting on Claude Sonnet 4.5 (RQ1). D\&C achieves 87.62\% F1, outperforming the monolithic baseline (52.14\% F1). The 54\%  precision improvement is most pronounced, indicating the monolithic approach suffers high false positive rates when reasoning over multiple error types simultaneously. Decomposition into focused sub-tasks with reduced context transforms production-unviable precision (43.38\%) into deployment-suitable performance (97.22\%).

\begin{table}[t]
\centering
\scriptsize
\setlength{\tabcolsep}{1pt}
\begin{tabular}{@{}cccccc@{}}
\toprule
\textbf{Model} & \textbf{Method} & \textbf{Precision} & \textbf{Recall} & \textbf{F1 score} \\
\midrule
\multirow{3}{*}{\makecell{Qwen3\\8B}} & Non FT & 0.7732 {\tiny($\pm$0.0543)} & 0.4536 {\tiny($\pm$0.0490)} & 0.5718 {\tiny($\pm$0.0452)} \\
& RBD & \textbf{0.9228} {\tiny($\pm$0.0311)} & 0.6744 {\tiny($\pm$0.0465)} & 0.7793 {\tiny($\pm$0.0346)} \\
& PTC & 0.9344 {\tiny($\pm$0.0271)} & \textbf{0.7667} {\tiny($\pm$0.0418)} & \textbf{0.8423} {\tiny($\pm$0.0290)} \\
\midrule
\multirow{3}{*}{\makecell{Mistral\\Nemo\\12B}} & Non FT & 0.5599 {\tiny($\pm$0.0451)} & 0.6104 {\tiny($\pm$0.0484)} & 0.5840 {\tiny($\pm$0.0393)} \\
& RBD & \textbf{0.9364} {\tiny($\pm$0.0285)} & 0.6795 {\tiny($\pm$0.0463)} & 0.7875 {\tiny($\pm$0.0340)} \\
& PTC & 0.9066 {\tiny($\pm$0.0315)} & \textbf{0.7718} {\tiny($\pm$0.0418)} & \textbf{0.8338} {\tiny($\pm$0.0295)} \\
\midrule
\multirow{3}{*}{\makecell{Qwen3\\14B}} & Non FT & 0.8607 {\tiny($\pm$0.0409)} & 0.7197 {\tiny($\pm$0.0446)} & 0.7839 {\tiny($\pm$0.0341)} \\
& RBD & \textbf{0.9386} {\tiny($\pm$0.0303)} & 0.7244 {\tiny($\pm$0.0422)} & 0.8176 {\tiny($\pm$0.0297)} \\
& PTC & 0.9112 {\tiny($\pm$0.0303)} & \textbf{0.7897} {\tiny($\pm$0.0408)} & \textbf{0.8462} {\tiny($\pm$0.0282)} \\
\midrule
\multirow{3}{*}{\makecell{Mistral\\Small\\24B}} & Non FT & 0.8424 {\tiny($\pm$0.0489)} & 0.5463 {\tiny($\pm$0.0495)} & 0.6628 {\tiny($\pm$0.0424)} \\
& RBD & \textbf{0.9192} {\tiny($\pm$0.0347)} & 0.7030 {\tiny($\pm$0.0444)} & 0.7967 {\tiny($\pm$0.0321)} \\
& PTC & 0.9135 {\tiny($\pm$0.0310)} & \textbf{0.7308} {\tiny($\pm$0.0440)} & \textbf{0.8120} {\tiny($\pm$0.0314)} \\
\midrule
\multirow{3}{*}{\makecell{Qwen3\\32B}} & Non FT & 0.8897 {\tiny($\pm$0.0441)} & 0.5558 {\tiny($\pm$0.0495)} & 0.6842 {\tiny($\pm$0.0415)} \\
& RBD & \textbf{0.9343} {\tiny($\pm$0.0291)} & 0.7434 {\tiny($\pm$0.0394)} & 0.8280 {\tiny($\pm$0.0277)} \\
& PTC & 0.9157 {\tiny($\pm$0.0294)} & \textbf{0.8077} {\tiny($\pm$0.0392)} & \textbf{0.8583} {\tiny($\pm$0.0271)} \\
\bottomrule
\end{tabular}
\caption{Performance comparison of non-finetuned , RBD, and PTC across five LLMs from the Qwen and Mistral families. PTC achieves the highest F1 score across all architectures.}
\label{tab:finetuning_results}
\end{table}

\paragraph{PTC vs baselines.}
Table~\ref{tab:finetuning_results} compares non-finetuned, RBD, and PTC models across five architectures (RQ2, RQ3). PTC consistently achieves the highest F1, with improvements over RBD ranging from 1.53\% (Mistral Small 24B) to 6.30\% (Qwen3 8B). PTC improvements are consistent across both Qwen and Mistral families, and the 8B model (84.23\% F1) matches the 14B variant (84.62\%), suggesting finetuning normalizes capacity differences. Both finetuning approaches substantially outperform non-finetuned baselines.

Results on synthetic data (Table~\ref{tab:synthetic_results}) confirm the same trends with higher absolute scores: PTC achieves 96.62\% F1 on Qwen3 8B versus 94.59\% for RBD. The absolute difference between real and synthetic performance reflects the inherent complexity of real product data. 
Real product data may contain cases where visually distinct products share near-identical names (e.g., \textit{Red1} and \textit{Red}) but represent genuinely different color variants within the same family. As the synthetic pipeline is text-only, it does not generate these ambiguous cases, yielding cleaner decision boundaries and higher recall.

\begin{table}[t]
\centering
\scriptsize
\setlength{\tabcolsep}{1pt}
\begin{tabular}{@{}cccccc@{}}
\toprule
\textbf{Model} & \textbf{Method} & \textbf{Precision} & \textbf{Recall} & \textbf{F1 score} \\
\midrule
\multirow{3}{*}{\makecell{Qwen3\\8B}} & Non FT & 0.9869 {\tiny($\pm$0.0041)} & 0.6434 {\tiny($\pm$0.0140)} & 0.7789 {\tiny($\pm$0.0104)} \\
& RBD & \textbf{0.9899} {\tiny($\pm$0.0030)} & 0.9055 {\tiny($\pm$0.0084)} & 0.9459 {\tiny($\pm$0.0048)} \\
& PTC & 0.9847 {\tiny($\pm$0.0036)} & \textbf{0.9483} {\tiny($\pm$0.0064)} & \textbf{0.9662} {\tiny($\pm$0.0038)} \\
\midrule
\multirow{3}{*}{\makecell{Mistral\\Nemo\\12B}} & Non FT & 0.9254 {\tiny($\pm$0.0102)} & 0.5112 {\tiny($\pm$0.0146)} & 0.6586 {\tiny($\pm$0.0128)} \\
& RBD & \textbf{0.9931} {\tiny($\pm$0.0032)} & 0.5708 {\tiny($\pm$0.0142)} & 0.7249 {\tiny($\pm$0.0115)} \\
& PTC & 0.9862 {\tiny($\pm$0.0035)} & \textbf{0.9531} {\tiny($\pm$0.0061)} & \textbf{0.9693} {\tiny($\pm$0.0036)} \\
\midrule
\multirow{3}{*}{\makecell{Qwen3\\14B}} & Non FT & \textbf{0.9842} {\tiny($\pm$0.0043)} & 0.6955 {\tiny($\pm$0.0135)} & 0.8151 {\tiny($\pm$0.0095)} \\
& RBD & 0.9694 {\tiny($\pm$0.0051)} & 0.9290 {\tiny($\pm$0.0075)} & 0.9487 {\tiny($\pm$0.0047)} \\
& PTC & 0.9836 {\tiny($\pm$0.0038)} & \textbf{0.9340} {\tiny($\pm$0.0072)} & \textbf{0.9582} {\tiny($\pm$0.0043)} \\
\midrule
\multirow{3}{*}{\makecell{Mistral\\Small\\24B}} & Non FT & \textbf{0.9841} {\tiny($\pm$0.0042)} & 0.7470 {\tiny($\pm$0.0127)} & 0.8493 {\tiny($\pm$0.0084)} \\
& RBD & 0.9760 {\tiny($\pm$0.0045)} & 0.9524 {\tiny($\pm$0.0061)} & 0.9640 {\tiny($\pm$0.0038)} \\
& PTC & 0.9790 {\tiny($\pm$0.0041)} & \textbf{0.9531} {\tiny($\pm$0.0060)} & \textbf{0.9659} {\tiny($\pm$0.0037)} \\
\bottomrule
\end{tabular}
\caption{Performance comparison of non-finetuned, RBD, and PTC on synthetic data across four LLMs from the Qwen and Mistral families. PTC achieves the highest F1 score for every model.}
\label{tab:synthetic_results}
\end{table}

\paragraph{Per task analysis.} Table~\ref{tab:per_error_results_qwen8b_only} shows PTC's gains are largest on the most challenging tasks. On duplicates, where the baseline fails (20.01\% F1), PTC achieves 77.84\% (a 17\% improvement over RBD). PTC also improves theme defects (5\% over RBD) and overstuffing (4\%). For unit systems, where RBD already performs well, gains are more modest (3\%). This pattern suggests PTC particularly benefits tasks requiring nuanced contextual reasoning. Full per task results across all models are given in Appendix ~\ref{sec:per_error_breakdown}.

\begin{table}[t]
\centering
\scriptsize
\setlength{\tabcolsep}{5pt}
\begin{tabular}{@{}lcccc@{}}
\toprule
\textbf{Method} & \textbf{Overstuffing} & \textbf{Unit sys.} & \textbf{Theme} & \textbf{Duplicates} \\
\midrule
Non FT & 0.6403 & 0.4348 & 0.4724 & 0.2001 \\
RBD & 0.7852 & 0.7619 & 0.7500 & 0.6065 \\
PTC & \textbf{0.8347} & \textbf{0.7925} & \textbf{0.8041} & \textbf{0.7784} \\
\bottomrule
\end{tabular}
\caption{Per task F1 on Qwen3 8B. 95\% bootstrap MoE range from $\pm$0.03 to $\pm$0.13. Full results with CI across all models in \S ~\ref{sec:per_error_breakdown}.}
\label{tab:per_error_results_qwen8b_only}
\end{table}

\begin{figure}[t]
    \centering
    \includegraphics[width=\columnwidth]{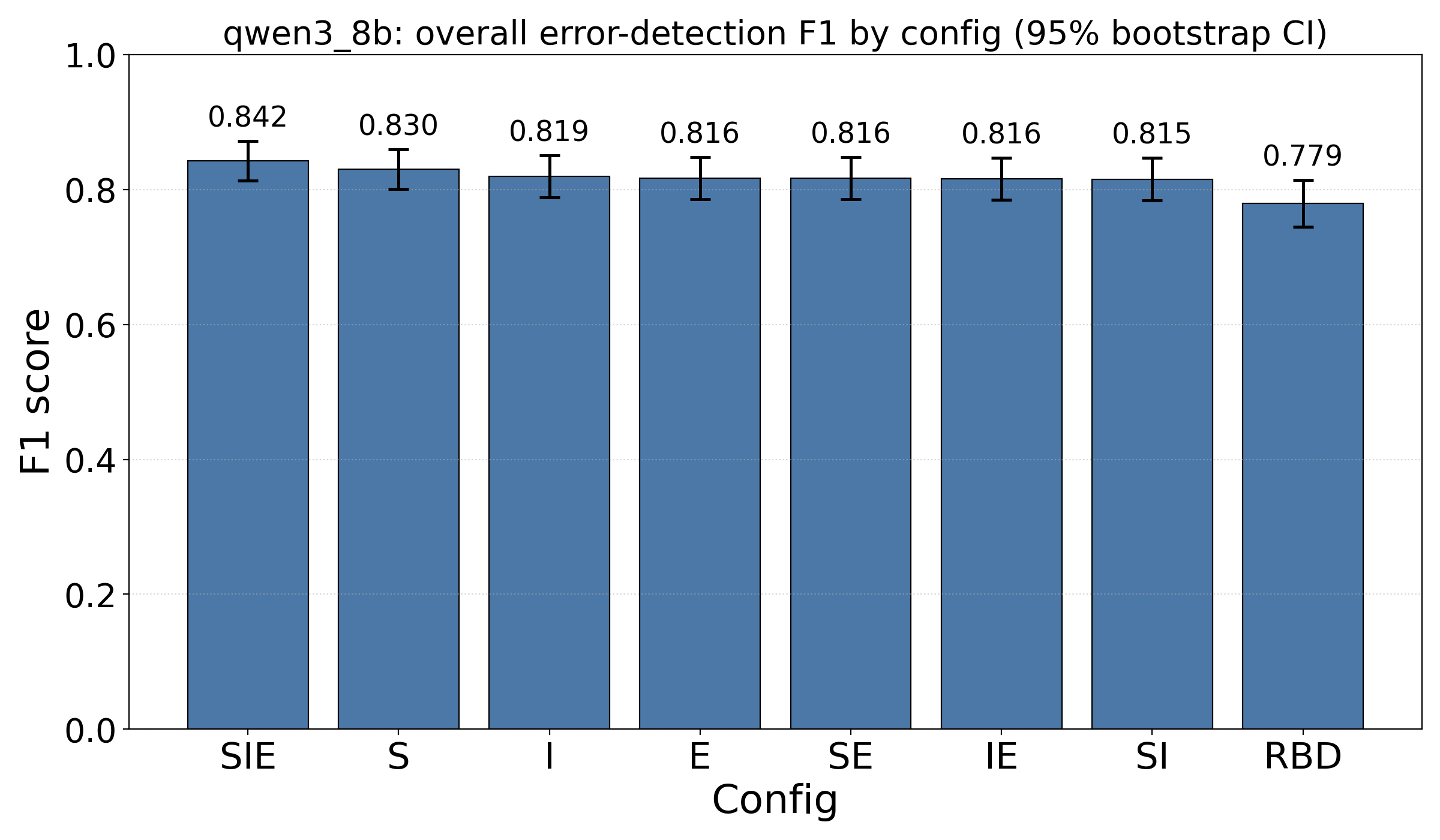}
    \caption{F1 scores across 7 positional configurations on Qwen3 8B with 95\% bootstrap CI. S = start of prompt, I = before input, E = end of prompt. The optimal configuration (SIE) places task tokens at the start, input, and after the end of the prompt, achieving 84.2\% F1. All PTC configurations outperform RBD (77.9\%).}
    \label{fig:positional_ablation}
\end{figure}

\begin{table*}[t]
\centering
\scriptsize
\setlength{\tabcolsep}{3pt}
\renewcommand{\arraystretch}{1.2}
\begin{tabularx}{\textwidth}{@{}p{0.8cm}p{0.6cm}X p{0.7cm}X p{0.7cm}X@{}}
\toprule
\textbf{Category} & \textbf{Attr.} & \textbf{Attribute values} & \textbf{RBD} & \textbf{RBD reason} & \textbf{PTC} & \textbf{PTC reason} \\
\midrule
Bed & color & Beige, Black, Dark Gray, Dark Grey, Light Gray, Light Grey, Navy, Olive Green, Pink, Sage, White, \ldots \emph{(27 values total)}
& correct
& All values are genuinely different color variations. Similar spellings like `Dark Gray' and `Dark Grey' represent standardized variants, not duplicates.
& egregious error
& `Dark Gray' and `Dark Grey' are present, differing only by the spelling of `Gray'/`Grey', meeting the duplicate criteria for spelling differences. \\
\bottomrule
\end{tabularx}
\caption{Qualitative duplicates example on Qwen3 8B where RBD labels the variation family correct but PTC detects the duplicate. The duplicate differs only by spelling (\texttt{Dark Gray}/\texttt{Dark Grey}). RBD treats these as intentional variants, while PTC's explicit task conditioning flags them as duplicates.}
\label{tab:duplicates_qualitative_small}
\end{table*}

\paragraph{Positional ablation.}
We evaluate 7 positional configurations on Qwen3 8B (Fig. ~\ref{fig:positional_ablation}, RQ4). All PTC configurations outperform RBD, confirming that explicit task conditioning is beneficial regardless of placement. The best configuration is SIE (84.2\% F1), placing task tokens at the three functional transition points: before the introduction (start), before the input data, and after the output (end). Among single positions, placement before the introduction (S: 83.0\%) outperforms others (I: 81.9\%, E: 81.6\%), aligning with the intuition that early conditioning establishes task-appropriate representations. Most multi-position configurations do not improve over the best single position; SIE is the exception, combining early priming (S) with reinforcement at the two boundaries where the model shifts processing mode, from instructions to data and from context to generation. 

\paragraph{Qualitative analysis: PTC vs RBD.}
Table~\ref{tab:duplicates_qualitative_small} illustrates the mechanism behind PTC's recall improvement over RBD. In this example, both models encounter a 27-value color list containing \textit{Dark Gray} and \textit{Dark Grey}, a clear spelling duplicate. RBD's reasoning explicitly identifies the pair but rationalizes it away, concluding they \textit{represent standardized variants, not duplicates.} PTC flags the pair as meeting the duplicate criteria for spelling differences. 
We hypothesize that the task token placed immediately before the input re-anchors attention at the point where the model transitions from instructions to data, PTC retains the focus on task, unlike RBD which defers to the majority class. 
This may explain PTC's systematic recall advantage over RBD (Table~\ref{tab:finetuning_results}): both models detect the same evidence, but PTC maintains sufficient task conditioning to act on it.

\paragraph{Cost analysis.} Although divide-and-conquer issues up to four LLM calls per attribute, its total token cost is comparable to a single monolithic call because each sub-task uses a much shorter, task-focused prompt. On the human-annotated test set, monolithic prompting costs approximately \$59 at public pricing (mean input 7000 tokens, output 900 tokens), versus \$63 for the D\&C solution (mean input 2500 tokens, output 150 tokens per call). Thus the accuracy gains of decomposition come at negligible additional teacher cost. For deployment, distillation yields the primary savings: the same D\&C workload on Mistral Nemo 12B with PTC completes in approximately 2 minutes on 8$\times$ H100 GPUs, costing approximately \$1.17 based on public instance pricing (\$35/hr) versus \$63 for Claude Sonnet 4.5, a 98\% cost reduction. The public pricing costs are at the time of writing.

\paragraph{Statistical significance.} We performed a statistical evaluation of the results, showing that the performance gains are statistically significant in the vast majority of cases. We assessed the significance using a paired subsampled bootstrap test with 5,000 class-balanced 80\% draws without replacement, applying a plus-one correction. The p values are shown in \S\ref{sec:statistical_significance}.

\section{Conclusion}
\label{sec:conclusion}

Our divide-and-conquer strategy decomposes monolithic detection into focused sub-tasks, improving F1 from 52.14\% to 87.62\% when using frontier LLM. The Positional Task Conditioning approach is able to distill this capability into smaller LLMs through explicit task conditioning, outperforming Rationale-Based Distillation across  five evaluated models (statistically significant on four out of five) achieving within 1.79\% F1 of the frontier model at upto 98\% lower inference cost. Future work should explore PTC in other product catalog quality tasks such as attribute extraction or entity matching to improve classifier quality.

\section{Limitations}
\label{sec:limitations}


First, our evaluation is conducted on a single domain. While consistent improvements across model families and scales suggest generalizability, we have not validated PTC on other multi-label classification tasks or alternative domains. 

Second, PTC successfully consolidates four sub-tasks into a single model, but the scalability of task conditioning to larger numbers of tasks remains unexplored. Whether a single model can maintain task discrimination as the number of tasks grows is an open question we leave to future work.

Third, we do not evaluate the impact of training data size through ablation studies. Although absolute performance likely varies with dataset scale, our comparative analysis remains robust because all methods were evaluated using identical data and hyperparameters. Consequently, whether PTC enhances sample efficiency relative to RBD is a question we leave to future research.

Finally, a notable limitation is the proprietary nature of our dataset. To mitigate this concern, we provide a thorough description of the quality assessment tasks alongside a detailed statistical breakdown of the dataset's composition.

\bibliography{custom}

@inproceedings{schmidts2020catalog,
  title={Catalog Integration of Low-quality Product Data by Attribute Label Ranking.},
  author={Schmidts, Oliver and Kraft, Bodo and Winkens, Marvin and Z{\"u}ndorf, Albert},
  booktitle={DATA},
  pages={90--101},
  year={2020}
}

@article{li2020deep,
  title={Deep entity matching with pre-trained language models},
  author={Li, Yuliang and Li, Jinfeng and Suhara, Yoshihiko and Doan, AnHai and Tan, Wang-Chiew},
  journal={arXiv preprint arXiv:2004.00584},
  year={2020}
}

@misc{negri2025attributeawarecontrolledproductgeneration,
      title={Attribute-Aware Controlled Product Generation with LLMs for E-commerce}, 
      author={Virginia Negri and Víctor Martínez Gómez and Sergio A. Balanya and Subburam Rajaram},
      year={2025},
      eprint={2601.04200},
      archivePrefix={arXiv},
      primaryClass={cs.CL},
      url={https://arxiv.org/abs/2601.04200}, 
}

@inproceedings{satyadharma2025auto,
  title={Auto prompting without training labels: An LLM cascade for product quality assessment in e-commerce catalogs},
  author={Satyadharma, Soham and Sheikholeslami, Fatemeh and Kaul, Swati and Batur, Aziz Umit and Khan, Suleiman A},
  booktitle={Proceedings of the 2025 Conference on Empirical Methods in Natural Language Processing: Industry Track},
  pages={937--953},
  year={2025}
}

@article{nikolakopoulos2023sage,
  title={Sage: Structured attribute value generation for billion-scale product catalogs},
  author={Nikolakopoulos, Athanasios N and Kaul, Swati and Gade, Siva Karthik and Dubrov, Bella and Batur, Umit and Khan, Suleiman Ali},
  journal={arXiv preprint arXiv:2309.05920},
  year={2023}
}

@inproceedings{wang2020automatic,
  title={Automatic validation of textual attribute values in e-commerce catalog by learning with limited labeled data},
  author={Wang, Yaqing and Xu, Yifan Ethan and Li, Xian and Dong, Xin Luna and Gao, Jing},
  booktitle={Proceedings of the 26th ACM SIGKDD International Conference on Knowledge Discovery \& Data Mining},
  pages={2533--2541},
  year={2020}
}

@book{ilyas2019data,
  title={Data cleaning},
  author={Ilyas, Ihab F and Chu, Xu},
  year={2019},
  publisher={Morgan \& Claypool}
}

@article{raffel2020exploring,
  title={Exploring the limits of transfer learning with a unified text-to-text transformer},
  author={Raffel, Colin and Shazeer, Noam and Roberts, Adam and Lee, Katherine and Narang, Sharan and Matena, Michael and Zhou, Yanqi and Li, Wei and Liu, Peter J},
  journal={Journal of machine learning research},
  volume={21},
  number={140},
  pages={1--67},
  year={2020}
}

@article{xu2015will,
  title={Will video be the next generation of e-commerce product reviews? Presentation format and the role of product type},
  author={Xu, Pei and Chen, Liang and Santhanam, Radhika},
  journal={Decision Support Systems},
  volume={73},
  pages={85--96},
  year={2015},
  publisher={Elsevier}
}

@article{liu2024lost,
  title={Lost in the middle: How language models use long contexts},
  author={Liu, Nelson F and Lin, Kevin and Hewitt, John and Paranjape, Ashwin and Bevilacqua, Michele and Petroni, Fabio and Liang, Percy},
  journal={Transactions of the Association for Computational Linguistics},
  volume={12},
  pages={157--173},
  year={2024}
}

@article{kou2025rethinking,
  title={Rethinking Toxicity Evaluation in Large Language Models: A Multi-Label Perspective},
  author={Kou, Zhiqiang and Chen, Junyang and Cai, Xin-Qiang and Xie, Ming-Kun and Liu, Biao and Wang, Changwei and Feng, Lei and Jia, Yuheng and Niu, Gang and Sugiyama, Masashi and others},
  journal={arXiv preprint arXiv:2510.15007},
  year={2025}
}

@misc{ortego2025largelanguagemodelsmeet,
      title={Large Language Models Meet Extreme Multi-label Classification: Scaling and Multi-modal Framework}, 
      author={Diego Ortego and Marlon Rodríguez and Mario Almagro and Kunal Dahiya and David Jiménez and Juan C. SanMiguel},
      year={2025},
      eprint={2511.13189},
      archivePrefix={arXiv},
      primaryClass={cs.CV},
      url={https://arxiv.org/abs/2511.13189}, 
}

@inproceedings{tabatabaei2025can,
  title={Can large language models serve as effective classifiers for hierarchical multi-label classification of scientific documents at industrial scale?},
  author={Tabatabaei, Seyed Amin and Fancher, Sarah and Parsons, Michael and Askari, Arian},
  booktitle={Proceedings of the 31st International Conference on Computational Linguistics: Industry Track},
  pages={163--174},
  year={2025}
}

@article{khot2022decomposed,
  title={Decomposed prompting: A modular approach for solving complex tasks},
  author={Khot, Tushar and Trivedi, Harsh and Finlayson, Matthew and Fu, Yao and Richardson, Kyle and Clark, Peter and Sabharwal, Ashish},
  journal={arXiv preprint arXiv:2210.02406},
  year={2022}
}

@article{zhou2022least,
  title={Least-to-most prompting enables complex reasoning in large language models},
  author={Zhou, Denny and Sch{\"a}rli, Nathanael and Hou, Le and Wei, Jason and Scales, Nathan and Wang, Xuezhi and Schuurmans, Dale and Cui, Claire and Bousquet, Olivier and Le, Quoc and others},
  journal={arXiv preprint arXiv:2205.10625},
  year={2022}
}

@article{hinton2015distilling,
  title={Distilling the knowledge in a neural network},
  author={Hinton, Geoffrey and Vinyals, Oriol and Dean, Jeff},
  journal={arXiv preprint arXiv:1503.02531},
  year={2015}
}

@article{gou2021knowledge,
  title={Knowledge distillation: A survey},
  author={Gou, Jianping and Yu, Baosheng and Maybank, Stephen J and Tao, Dacheng},
  journal={International journal of computer vision},
  volume={129},
  number={6},
  pages={1789--1819},
  year={2021},
  publisher={Springer}
}

@article{caruana1997multitask,
  title={Multitask learning},
  author={Caruana, Rich},
  journal={Machine learning},
  volume={28},
  number={1},
  pages={41--75},
  year={1997},
  publisher={Springer}
}

@article{yang2025qwen3,
  title={Qwen3 technical report},
  author={Yang, An and Li, Anfeng and Yang, Baosong and Zhang, Beichen and Hui, Binyuan and Zheng, Bo and Yu, Bowen and Gao, Chang and Huang, Chengen and Lv, Chenxu and others},
  journal={arXiv preprint arXiv:2505.09388},
  year={2025}
}

@misc{mistralnemo2024,
      title={Mistral NeMo},
      author={{Mistral AI} and {NVIDIA}},
      year={2024},
      howpublished={\url{https://mistral.ai/news/mistral-nemo/}},
      note={Released July 2024}
}

@misc{mistralsmall24b,
      title={Mistral Small: Optimized Portability and Performance},
      author={{Mistral AI}},
      year={2024},
      howpublished={\url{https://mistral.ai/news/mistral-small-3/}},
      note={24B Parameter Model}
}

@article{dettmers2023qlora,
  title={Qlora: Efficient finetuning of quantized llms},
  author={Dettmers, Tim and Pagnoni, Artidoro and Holtzman, Ari and Zettlemoyer, Luke},
  journal={Advances in neural information processing systems},
  volume={36},
  pages={10088--10115},
  year={2023}
}

@inproceedings{devlin2019bert,
  title={Bert: Pre-training of deep bidirectional transformers for language understanding},
  author={Devlin, Jacob and Chang, Ming-Wei and Lee, Kenton and Toutanova, Kristina},
  booktitle={Proceedings of the 2019 conference of the North American chapter of the association for computational linguistics: human language technologies, volume 1 (long and short papers)},
  pages={4171--4186},
  year={2019}
}

@article{liu2019roberta,
  title={Roberta: A robustly optimized bert pretraining approach},
  author={Liu, Yinhan and Ott, Myle and Goyal, Naman and Du, Jingfei and Joshi, Mandar and Chen, Danqi and Levy, Omer and Lewis, Mike and Zettlemoyer, Luke and Stoyanov, Veselin},
  journal={arXiv preprint arXiv:1907.11692},
  year={2019}
}

@inproceedings{niraula2024multi,
  title={Multi-label classification with generative large language models},
  author={Niraula, Nobal and Ayhan, Samet and Chidambaram, Balaguruna and Whyatt, Daniel},
  booktitle={2024 AIAA DATC/IEEE 43rd Digital Avionics Systems Conference (DASC)},
  pages={1--7},
  year={2024},
  organization={IEEE}
}

@article{ma2025large,
  title={Large Language Models Do Multi-Label Classification Differently},
  author={Ma, Marcus and Chochlakis, Georgios and Pandiyan, Niyantha Maruthu and Thomason, Jesse and Narayanan, Shrikanth},
  journal={arXiv preprint arXiv:2505.17510},
  year={2025}
}

@inproceedings{zhou2024quest,
  title={Quest: Efficient extreme multi-label text classification with large language models on commodity hardware},
  author={Zhou, Chuang and Dong, Junnan and Huang, Xiao and Liu, Zirui and Zhou, Kaixiong and Xu, Zhaozhuo},
  booktitle={Findings of the Association for Computational Linguistics: EMNLP 2024},
  pages={3929--3940},
  year={2024}
}

@article{wei2022chain,
  title={Chain-of-thought prompting elicits reasoning in large language models},
  author={Wei, Jason and Wang, Xuezhi and Schuurmans, Dale and Bosma, Maarten and Xia, Fei and Chi, Ed and Le, Quoc V and Zhou, Denny and others},
  journal={Advances in neural information processing systems},
  volume={35},
  pages={24824--24837},
  year={2022}
}

@article{kojima2022large,
  title={Large language models are zero-shot reasoners},
  author={Kojima, Takeshi and Gu, Shixiang Shane and Reid, Machel and Matsuo, Yutaka and Iwasawa, Yusuke},
  journal={Advances in neural information processing systems},
  volume={35},
  pages={22199--22213},
  year={2022}
}

@article{wang2023scott,
  title={Scott: Self-consistent chain-of-thought distillation},
  author={Wang, Peifeng and Wang, Zhengyang and Li, Zheng and Gao, Yifan and Yin, Bing and Ren, Xiang},
  journal={arXiv preprint arXiv:2305.01879},
  year={2023}
}

@inproceedings{magister2023teaching,
  title={Teaching small language models to reason},
  author={Magister, Lucie Charlotte and Mallinson, Jonathan and Adamek, Jakub and Malmi, Eric and Severyn, Aliaksei},
  booktitle={Proceedings of the 61st annual meeting of the association for computational linguistics (volume 2: short papers)},
  pages={1773--1781},
  year={2023}
}

@inproceedings{hsieh2023distilling,
  title={Distilling step-by-step! outperforming larger language models with less training data and smaller model sizes},
  author={Hsieh, Cheng-Yu and Li, Chun-Liang and Yeh, Chih-Kuan and Nakhost, Hootan and Fujii, Yasuhisa and Ratner, Alex and Krishna, Ranjay and Lee, Chen-Yu and Pfister, Tomas},
  booktitle={Findings of the Association for Computational Linguistics: ACL 2023},
  pages={8003--8017},
  year={2023}
}

@article{mukherjee2023orca,
  title={Orca: Progressive learning from complex explanation traces of gpt-4},
  author={Mukherjee, Subhabrata and Mitra, Arindam and Jawahar, Ganesh and Agarwal, Sahaj and Palangi, Hamid and Awadallah, Ahmed},
  journal={arXiv preprint arXiv:2306.02707},
  year={2023}
}

@inproceedings{ho2023large,
  title={Large language models are reasoning teachers},
  author={Ho, Namgyu and Schmid, Laura and Yun, Se-Young},
  booktitle={Proceedings of the 61st annual meeting of the association for computational linguistics (volume 1: long papers)},
  pages={14852--14882},
  year={2023}
}

@article{maragheh2023llm,
  title={Llm-based aspect augmentations for recommendation systems},
  author={Maragheh, Reza Yousefi and Morishetti, Lalitesh and Giahi, Ramin and Nag, Kaushiki and Xu, Jianpeng and Cho, Jason and Korpeoglu, Evren and Kumar, Sushant and Achan, Kannan},
  year={2023}
}

@article{chen2023knowledge,
  title={Knowledge graph completion models are few-shot learners: An empirical study of relation labeling in e-commerce with llms},
  author={Chen, Jiao and Ma, Luyi and Li, Xiaohan and Thakurdesai, Nikhil and Xu, Jianpeng and Cho, Jason HD and Nag, Kaushiki and Korpeoglu, Evren and Kumar, Sushant and Achan, Kannan},
  journal={arXiv preprint arXiv:2305.09858},
  year={2023}
}

@article{wang2024leveraging,
  title={Leveraging Large Language Models for Context-Aware Product Discovery in E-commerce Search Systems},
  author={Wang, Gaike and Ni, Xin and Shen, Qi and Yang, Mingxuan},
  journal={Journal of Knowledge Learning and Science Technology ISSN: 2959-6386 (online)},
  volume={3},
  number={4},
  pages={300--312},
  year={2024}
}

@misc{rokon2026enhancementecommercesponsoredsearch,
      title={Enhancement of E-commerce Sponsored Search Relevancy with LLM}, 
      author={Md Omar Faruk Rokon and Andrei Simion and Weizhi Du and Musen Wen and Hong Yao and Kuang-chih Lee},
      year={2026},
      eprint={2607.03886},
      archivePrefix={arXiv},
      primaryClass={cs.IR},
      url={https://arxiv.org/abs/2607.03886}, 
}

@article{herrero2024learning,
  title={Learning variant product relationship and variation attributes from e-commerce website structures},
  author={Herrero-Vidal, Pedro and Chen, You-Lin and Liu, Cris and Sen, Prithviraj and Wang, Lichao},
  journal={arXiv preprint arXiv:2410.02779},
  year={2024}
}

@inproceedings{cheng2024commerce,
  title={E-commerce product categorization with LLM-based dual-expert classification paradigm},
  author={Cheng, Zhu and Zhang, Wen and Chou, Chih-Chi and Jau, You-Yi and Pathak, Archita and Gao, Peng and Batur, Umit},
  booktitle={Proceedings of the 1st Workshop on Customizable NLP: Progress and Challenges in Customizing NLP for a Domain, Application, Group, or Individual (CustomNLP4U)},
  pages={294--304},
  year={2024}
}

@article{kathiriya2023optimizing,
  title={Optimizing ECommerce Listing: LLM Based Description and Keyword Generation from Multimodal Data},
  author={Kathiriya, Satish and Mullapudi, Mahidhar and Karangara, Rajath},
  journal={Int. J. Sci. Res.(IJSR)},
  volume={12},
  pages={2123--2130},
  year={2023}
}

@inproceedings{baumann2024using,
  author    = {Nick Baumann and Alexander Brinkmann and Christian Bizer},
  title     = {Using LLMs for the Extraction and Normalization of Product Attribute Values},
  booktitle = {Proceedings of the 28th European Conference on Advances in Databases and Information Systems (ADBIS 2024)},
  year      = {2024},
}

@inproceedings{zhang2025leveraging,
  title={Leveraging product catalog patterns for multilingual e-commerce product attribute prediction},
  author={Zhang, Bryan and Khan, Suleiman A and Walter, Stephan},
  booktitle={Proceedings of the 2025 Conference on Empirical Methods in Natural Language Processing: Industry Track},
  pages={267--275},
  year={2025}
}

@inproceedings{trabelsi2025matters,
  title={What matters when building vision language models for product image analysis?},
  author={Trabelsi, Ameni and Zontak, Maria and Qian, Yiming and Jackson, Brian and Khan, Suleiman and Batur, Umit},
  booktitle={Proceedings of the Winter Conference on Applications of Computer Vision},
  pages={1372--1381},
  year={2025}
}

@software{axolotl,
  title = {Axolotl: Open Source LLM Post-Training},
  author = {{Axolotl maintainers and contributors}},
  url = {https://github.com/axolotl-ai-cloud/axolotl},
  license = {Apache-2.0},
  year = {2023}
}

@inproceedings{fu2023specializing,
  title={Specializing smaller language models towards multi-step reasoning},
  author={Fu, Yao and Peng, Hao and Ou, Litu and Sabharwal, Ashish and Khot, Tushar},
  booktitle={International Conference on Machine Learning},
  pages={10421--10430},
  year={2023},
  organization={PMLR}
}

@inproceedings{kwon2023efficient,
  title={Efficient Memory Management for Large Language Model Serving with PagedAttention},
  author={Woosuk Kwon and Zhuohan Li and Siyuan Zhuang and Ying Sheng and Lianmin Zheng and Cody Hao Yu and Joseph E. Gonzalez and Hao Zhang and Ion Stoica},
  booktitle={Proceedings of the ACM SIGOPS 29th Symposium on Operating Systems Principles},
  year={2023}
}

@inproceedings{liu2019multi,
  title={Multi-task deep neural networks for natural language understanding},
  author={Liu, Xiaodong and He, Pengcheng and Chen, Weizhu and Gao, Jianfeng},
  booktitle={Proceedings of the 57th annual meeting of the association for computational linguistics},
  pages={4487--4496},
  year={2019}
}

@article{ruder2017overview,
  title={An overview of multi-task learning in deep neural networks},
  author={Ruder, Sebastian},
  journal={arXiv preprint arXiv:1706.05098},
  year={2017}
}

@inproceedings{mansoori2026improving,
  title={Improving Cascade Routing for Structured Attribute Generation with Heterogeneous Confidence},
  author={Mansoori, Fatemeh and Scarinci, Andrea and Aggarwal, Aditya and Khan, Suleiman A and Chandramouli, Ashwin},
  booktitle={AdaptFM: Resource-Adaptive Foundation Model Inference},
  year={2026}
}

@inproceedings{venkataraman2026framework,
  title={A Framework for Prompt Optimization and Translation Across Foundation Models},
  author={Venkataraman, Abhinav Shankaranarayanan and Nikolakopoulos, Athanasios and Kumaraswamy, Vishwanath and Zhang, Tao and Chander, Sarath and Saboo, Rohit and Khan, Suleiman A.},
  booktitle={ICLR 2026 Workshop on AI with Recursive Self-Improvement},
  year={2026},
  url={https://openreview.net/forum?id=gOTCn5uZeE}
}

@article{opsahl2024optimizing,
  title={Optimizing Instructions and Demonstrations for Multi-Stage Language Model Programs},
  author={Opsahl-Ong, Krista and Ryan, Michael J and Purtell, Josh and Broman, David and Potts, Christopher and Zaharia, Matei and Khattab, Omar},
  journal={arXiv preprint arXiv:2406.11695},
  year={2024}
}

@article{gao2025prompt,
  title={The Prompt Alchemist: Automated LLM-Tailored Prompt Optimization for Test Case Generation},
  author={Gao, Shuzheng and Wang, Chaozheng and Gao, Cuiyun and Jiao, Xiaoqian and Chong, Chun Yong and Gao, Shan and Lyu, Michael},
  journal={arXiv preprint arXiv:2501.01329},
  year={2025}
}






\clearpage

\appendix




\section{Prompt templates}
\label{sec:prompt_templates}

This section presents the prompt template structure used for PTC. Figure~\ref{fig:ptc_size_overstuffing} shows an example template for detecting overstuffing in the size attribute. All prompts follow a consistent structure across tasks, with task-specific content in each section.

The task identifier (e.g., \texttt{<TASK=OVERSTUFFING>}) is inserted at three structural boundaries as described in Sec.~\ref{sec:positional_task_conditioning}: (1) before the introduction, establishing task-appropriate attention patterns from the first layer of processing; (2) before the input data, re-anchoring task context at the transition from instructions to data; and (3) after the output specification, conditioning the autoregressive generation phase.

The \texttt{<introduction>} section defines the classification objective and provides context about the attribute under evaluation. The \texttt{<rules>} and \texttt{<examples>} sections collectively constitute the instructions defined in Sec.~\ref{sec:divide_and_conquer}: \texttt{<rules>} contains detailed criteria for each severity level (no error, non-egregious, and egregious), capturing domain-specific conventions and edge cases refined with subject matter experts, while \texttt{<examples>} provides 6-8 few-shot demonstrations per class to calibrate model outputs. The \texttt{<input>} section contains the attribute values to be evaluated, and the \texttt{<output>} section specifies the required response format, including fields for the identified pattern, rationale, and final prediction. This structured format ensures consistent, parseable outputs suitable for both distillation and production deployment.

\begin{figure*}[h!]
    \begin{tcolorbox}[colback=yellow!20, colframe=black, coltext=black, fontupper=\ttfamily]  
    <TASK=OVERSTUFFING> \\ \\ 
    <introduction> \\
    Classify the given list of product size attribute values into one of three categories: NOT OVERSTUFFED, CONSISTENTLY OVERSTUFFED, or INCONSISTENTLY OVERSTUFFED. \\
    The list contains values that appear in the size attribute field for product families. \\
    You must analyze whether these values contain only size information or include additional attributes that don't belong in a size field (overstuffing). \\
    </introduction> \\ \\
    
    <rules> \\
    This section defines the specific criteria for each classification category: \\
    NOT OVERSTUFFED: \\ 
    \#\#\# Rules for NOT OVERSTUFFED go here. \\ 
    CONSISTENTLY OVERSTUFFED \\
    \#\#\# Rules for CONSISTENTLY OVERSTUFFED go here. \\ 
    INCONSISTENTLY OVERSTUFFED \\
    \#\#\# Rules for INCONSISTENTLY OVERSTUFFED go here. \\
    </rules> \\ \\

    <examples> \\
    \#\#\# 6-8 examples of each class go here. \\
    </examples> \\ \\
    
    <TASK=OVERSTUFFING> \\
    <input> \\
    \#\#\# Input data goes here. \\
    </input> \\ \\

    <output> \\
    Output the results in the following output format. \\
    \#\#\# Output format here. \\
    </output> \\ \\
    <TASK=OVERSTUFFING>
    \end{tcolorbox}
    \caption{ Example prompt template for PTC. The task identifier \texttt{<TASK=OVERSTUFFING>} is prepended to certain areas of the prompt, providing explicit task conditioning. The template comprises five sections: introduction (task definition), rules (classification criteria for each severity level), input (family attribute values), examples (few-shot examples), and output (response format specification).}
    \label{fig:ptc_size_overstuffing}
\end{figure*}

\section{Error categorization}
\label{app:error_categorization}

We have eight error types organized into four tasks, each comprising an egregious (E) and non-egregious (NE) variant. Egregious errors directly impair customer decisions—for instance, duplicate variations force customers to determine whether \textit{XL} and \textit{Extra Large} are truly different options, while mixed unit systems (\textit{Large} vs.\ \textit{10}) prevent meaningful comparison. Non-egregious errors are quality concerns that do not immediately mislead but degrade the browsing experience over time.

The distinction between egregious and non-egregious within each task captures a key operational insight: consistency mitigates severity. Overstuffing that applies uniformly across all values (every color includes a pack count) is less harmful than inconsistent overstuffing (only some values include extraneous information), because consistent patterns remain interpretable even if suboptimal. Similarly, theme errors where all values belong to a different attribute (all size values are actually scents) indicate a systematic misconfiguration, whereas inconsistent theme errors (a mix of colors and quantities) suggest data corruption that actively confuses customers.

\section{Per task breakdown}
\label{sec:per_error_breakdown}

\subsection{Human-annotated data}

Table~\ref{tab:per_error_results} reports F1 scores for each task across all models and methods on the human-annotated test set. Performance varies substantially across tasks, revealing that some defect categories are inherently more challenging for LLMs to detect. Overstuffing proves the most accessible task, with baselines achieving 53.12\%-73.93\% F1 and finetuned models reaching 77.43\%-83.47\%. Unit systems and Theme represent intermediate-difficulty tasks where baselines are inconsistent (31.22\%-74.14\%) but finetuning yields reliable detection (71.58\%-83.05\%). Duplicates is the most challenging category: baselines catastrophically fail on most models (Qwen3 8B: 20.01\%, Mistral Nemo 12B: 9.93\%, Qwen3 32B: 12.28\%), suggesting that duplicate detection requires reasoning capabilities that emerge only through targeted finetuning. Both RBD and PTC dramatically improve duplicates detection, with PTC achieving 77.84\% on Qwen3 8B and 75.00\% on Mistral Small 24B, representing 289\% and 266\% relative improvements over their respective baselines.

PTC's advantage over RBD is most pronounced on the tasks where RBD leaves the largest gap. For duplicates, PTC outperforms RBD by 28\% on Qwen3 8B (77.84\% vs.\ 60.65\%) and 10\% on Mistral Small 24B (75.00\% vs.\ 68.06\%). A similar pattern emerges for theme defect detection, where PTC surpasses RBD by 7\% on Qwen3 8B (80.41\% vs.\ 75.00\%) and 7\% on Mistral Small 24B (82.69\% vs.\ 77.55\%). In contrast, for overstuffing, where RBD already achieves strong performance, PTC provides more modest gains. This pattern suggests that positional task conditioning particularly benefits error types requiring multi-step contextual reasoning (identifying semantic duplicates across products, detecting thematic inconsistencies), where sustained task identity throughout the prompt is critical for the model to maintain focus on the specific defect pattern.

\begin{table}[t]
\centering
\scriptsize
\setlength{\tabcolsep}{1pt}
\begin{tabular}{@{}cccccc@{}}
\toprule
\textbf{Model} & \textbf{Method} & \textbf{Overstuffing} & \textbf{Unit systems} & \textbf{Theme} & \textbf{Duplicates} \\
\midrule
\multirow{3}{*}{Qwen3 8B} & Non FT & \makecell{0.6403\\{\tiny$\pm$0.0503}} & \makecell{0.4348\\{\tiny$\pm$0.1294}} & \makecell{0.4724\\{\tiny$\pm$0.1073}} & \makecell{0.2001\\{\tiny$\pm$0.0326}} \\
& RBD & \makecell{0.7852\\{\tiny$\pm$0.0417}} & \makecell{0.7619\\{\tiny$\pm$0.0924}} & \makecell{0.7500\\{\tiny$\pm$0.0976}} & \makecell{0.6065\\{\tiny$\pm$0.0911}} \\
& PTC & \makecell{\textbf{0.8347}\\{\tiny$\pm$0.0365}} & \makecell{\textbf{0.7925}\\{\tiny$\pm$0.0853}} & \makecell{\textbf{0.8041}\\{\tiny$\pm$0.0878}} & \makecell{\textbf{0.7784}\\{\tiny$\pm$0.0671}} \\
\midrule
\multirow{3}{*}{Mistral Nemo 12B} & Non FT & \makecell{0.5312\\{\tiny$\pm$0.0469}} & \makecell{0.4737\\{\tiny$\pm$0.0987}} & \makecell{0.3122\\{\tiny$\pm$0.0772}} & \makecell{0.0993\\{\tiny$\pm$0.0684}} \\
& RBD & \makecell{0.7743\\{\tiny$\pm$0.0434}} & \makecell{0.8000\\{\tiny$\pm$0.0849}} & \makecell{0.7158\\{\tiny$\pm$0.1037}} & \makecell{0.6627\\{\tiny$\pm$0.0833}} \\
& PTC & \makecell{\textbf{0.8106}\\{\tiny$\pm$0.0380}} & \makecell{\textbf{0.8034}\\{\tiny$\pm$0.0800}} & \makecell{\textbf{0.7473}\\{\tiny$\pm$0.1016}} & \makecell{\textbf{0.7234}\\{\tiny$\pm$0.0728}} \\
\midrule
\multirow{3}{*}{Qwen3 14B} & Non FT & \makecell{0.7393\\{\tiny$\pm$0.0428}} & \makecell{0.7414\\{\tiny$\pm$0.0900}} & \makecell{0.5152\\{\tiny$\pm$0.1050}} & \makecell{0.6354\\{\tiny$\pm$0.0803}} \\
& RBD & \makecell{0.8085\\{\tiny$\pm$0.0394}} & \makecell{0.7652\\{\tiny$\pm$0.0866}} & \makecell{0.7273\\{\tiny$\pm$0.0946}} & \makecell{\textbf{0.7647}\\{\tiny$\pm$0.0650}} \\
& PTC & \makecell{\textbf{0.8159}\\{\tiny$\pm$0.0384}} & \makecell{\textbf{0.8305}\\{\tiny$\pm$0.0746}} & \makecell{\textbf{0.7600}\\{\tiny$\pm$0.0933}} & \makecell{0.7615\\{\tiny$\pm$0.0628}} \\
\midrule
\multirow{3}{*}{Mistral Small 24B} & Non FT & \makecell{0.6816\\{\tiny$\pm$0.0477}} & \makecell{0.6804\\{\tiny$\pm$0.1074}} & \makecell{0.5229\\{\tiny$\pm$0.0967}} & \makecell{0.2051\\{\tiny$\pm$0.0984}} \\
& RBD & \makecell{\textbf{0.8065}\\{\tiny$\pm$0.0383}} & \makecell{\textbf{0.8246}\\{\tiny$\pm$0.0768}} & \makecell{0.7755\\{\tiny$\pm$0.0927}} & \makecell{0.6806\\{\tiny$\pm$0.0757}} \\
& PTC & \makecell{0.7881\\{\tiny$\pm$0.0413}} & \makecell{0.8155\\{\tiny$\pm$0.0827}} & \makecell{\textbf{0.8269}\\{\tiny$\pm$0.0804}} & \makecell{\textbf{0.7500}\\{\tiny$\pm$0.0682}} \\
\midrule
\multirow{3}{*}{Qwen3 32B} & Non FT & \makecell{0.7381\\{\tiny$\pm$0.0438}} & \makecell{0.6731\\{\tiny$\pm$0.1046}} & \makecell{0.6260\\{\tiny$\pm$0.0972}} & \makecell{0.1228\\{\tiny$\pm$0.0839}} \\
& RBD & \makecell{0.8168\\{\tiny$\pm$0.0381}} & \makecell{0.8000\\{\tiny$\pm$0.0834}} & \makecell{0.7800\\{\tiny$\pm$0.0911}} & \makecell{0.8000\\{\tiny$\pm$0.0605}} \\
& PTC & \makecell{\textbf{0.8293}\\{\tiny$\pm$0.0362}} & \makecell{\textbf{0.8174}\\{\tiny$\pm$0.0781}} & \makecell{\textbf{0.7863}\\{\tiny$\pm$0.0827}} & \makecell{\textbf{0.8186}\\{\tiny$\pm$0.0563}} \\
\bottomrule
\end{tabular}
\caption{Per task F1 (with 95\% bootstrap margin of error) for non-finetuned, RBD, and PTC across five LLMs on human labeled data.}
\label{tab:per_error_results}
\end{table}

\subsection{Synthetic data}

Table~\ref{tab:synthetic_per_error_results} reports per task F1 on the synthetic test set across four models. Qwen3 32B was excluded from synthetic evaluation due to prohibitive computational cost. The same difficulty ordering observed in real data holds: duplicates remains the hardest task for baselines (32.39\%-45.99\%), while overstuffing is more accessible. However, absolute scores are substantially higher across the board, reflecting the cleaner decision boundaries of synthetic data (see Sec.~\ref{sec:results}).

PTC achieves the highest F1 on nearly every task and model combination. The most striking result is on duplicates, where PTC reaches 94.04\% on Qwen3 8B versus 85.04\% for RBD—a 11\% improvement—and 93.78\% on Mistral Nemo 12B versus 88.93\% for RBD (5\% improvement).  PTC's consistent improvements across all tasks and models on synthetic data confirm that positional task conditioning provides robust task discrimination independent of data complexity.

\begin{table}[t]
\centering
\scriptsize
\setlength{\tabcolsep}{1pt}
\begin{tabular}{@{}cccccc@{}}
\toprule
\textbf{Model} & \textbf{Method} & \textbf{Overstuffing} & \textbf{Unit systems} & \textbf{Theme} & \textbf{Duplicates} \\
\midrule
\multirow{3}{*}{Qwen3 8B} & Non FT & \makecell{0.8498\\{\tiny$\pm$0.0100}} & \makecell{0.6676\\{\tiny$\pm$0.0278}} & \makecell{0.8618\\{\tiny$\pm$0.0141}} & \makecell{0.3918\\{\tiny$\pm$0.0317}} \\
& RBD & \makecell{0.9649\\{\tiny$\pm$0.0046}} & \makecell{0.8589\\{\tiny$\pm$0.0169}} & \makecell{0.9568\\{\tiny$\pm$0.0081}} & \makecell{0.8504\\{\tiny$\pm$0.0157}} \\
& PTC & \makecell{\textbf{0.9680}\\{\tiny$\pm$0.0044}} & \makecell{\textbf{0.8671}\\{\tiny$\pm$0.0163}} & \makecell{\textbf{0.9582}\\{\tiny$\pm$0.0077}} & \makecell{\textbf{0.9404}\\{\tiny$\pm$0.0090}} \\
\midrule
\multirow{3}{*}{Mistral Nemo 12B} & Non FT & \makecell{0.6307\\{\tiny$\pm$0.0163}} & \makecell{0.5212\\{\tiny$\pm$0.0297}} & \makecell{0.5403\\{\tiny$\pm$0.0271}} & \makecell{0.3239\\{\tiny$\pm$0.0290}} \\
& RBD & \makecell{0.4544\\{\tiny$\pm$0.0188}} & \makecell{0.8422\\{\tiny$\pm$0.0180}} & \makecell{0.6194\\{\tiny$\pm$0.0234}} & \makecell{0.8893\\{\tiny$\pm$0.0131}} \\
& PTC & \makecell{\textbf{0.9694}\\{\tiny$\pm$0.0044}} & \makecell{\textbf{0.8789}\\{\tiny$\pm$0.0159}} & \makecell{\textbf{0.9606}\\{\tiny$\pm$0.0074}} & \makecell{\textbf{0.9378}\\{\tiny$\pm$0.0093}} \\
\midrule
\multirow{3}{*}{Qwen3 14B} & Non FT & \makecell{0.8745\\{\tiny$\pm$0.0090}} & \makecell{0.8300\\{\tiny$\pm$0.0187}} & \makecell{0.8836\\{\tiny$\pm$0.0118}} & \makecell{0.4521\\{\tiny$\pm$0.0292}} \\
& RBD & \makecell{0.9569\\{\tiny$\pm$0.0050}} & \makecell{0.8507\\{\tiny$\pm$0.0174}} & \makecell{0.9417\\{\tiny$\pm$0.0091}} & \makecell{0.9031\\{\tiny$\pm$0.0109}} \\
& PTC & \makecell{\textbf{0.9704}\\{\tiny$\pm$0.0042}} & \makecell{\textbf{0.8544}\\{\tiny$\pm$0.0175}} & \makecell{\textbf{0.9620}\\{\tiny$\pm$0.0073}} & \makecell{\textbf{0.9080}\\{\tiny$\pm$0.0114}} \\
\midrule
\multirow{3}{*}{Mistral Small 24B} & Non FT & \makecell{0.9199\\{\tiny$\pm$0.0071}} & \makecell{0.8580\\{\tiny$\pm$0.0179}} & \makecell{0.8875\\{\tiny$\pm$0.0120}} & \makecell{0.4599\\{\tiny$\pm$0.0291}} \\
& RBD & \makecell{\textbf{0.9644}\\{\tiny$\pm$0.0046}} & \makecell{0.8613\\{\tiny$\pm$0.0165}} & \makecell{0.9411\\{\tiny$\pm$0.0088}} & \makecell{0.9443\\{\tiny$\pm$0.0081}} \\
& PTC & \makecell{0.9637\\{\tiny$\pm$0.0047}} & \makecell{\textbf{0.8626}\\{\tiny$\pm$0.0167}} & \makecell{\textbf{0.9537}\\{\tiny$\pm$0.0080}} & \makecell{\textbf{0.9463}\\{\tiny$\pm$0.0083}} \\
\bottomrule
\end{tabular}
\caption{Per task F1 (with 95\% bootstrap margin of error) for non-finetuned, RBD, and PTC on synthetic data across four LLMs. Bold marks the best F1 per task within each model.}
\label{tab:synthetic_per_error_results}
\end{table}

\section{Qualitative examples}
\label{sec:qualitative_examples}
\begin{table*}[t]
\centering
\scriptsize
\setlength{\tabcolsep}{3pt}
\renewcommand{\arraystretch}{1.2}
\begin{tabularx}{\textwidth}{@{}p{1.0cm}p{0.7cm}X p{0.8cm}X p{0.9cm}X@{}}
\toprule
\textbf{Category} & \textbf{Attr.} & \textbf{Attribute values} & \textbf{RBD label} & \textbf{RBD reason} & \textbf{PTC label} & \textbf{PTC reason} \\
\midrule
Bed & color & Beige, Black, Black \& Light Grey, Dark \& Light Grey, Dark Gray, Dark Grey, Dark Grey \& Light Grey, Green \& Ivory, Green \& Light Grey, Greyish Blue, Greyish Purple \& Light Grey, Light Gray, Light Grey, Navy, Navy Blue \& Grey, Olive Green, Orange \& Light Grey, Pink, Pink \& Ivory, Purple, Red \& Black, Royal Blue \& Light Grey, Sage, Sage Green \& Light Grey, Teal Blue \& Ivory, White, Yellow \& Light Grey
& correct
& All color values in the list are genuinely different color variations. While there are similar base colors (e.g., `Dark Gray' and `Dark Grey', `Light Gray' and `Light Grey'), these represent different standardized spellings of gray shades rather than duplicates, as they are commonly accepted as distinct color variations in product listings. There are no casing, punctuation, spacing, or spelling duplicates where both the correct and incorrect versions exist with no other differences.
& egregious error
& `Dark Gray' and `Dark Grey' appear in the list. These two values represent the same color with only spelling differences between `Gray' and `Grey'. Both variations are present and are otherwise completely identical with no additional descriptive words. This meets the duplicate criteria for spelling differences. All other values represent genuinely different color variations (distinct base colors, shades, and combinations). \\
\midrule
Shoes & color & Angora/Light Mango, Beige/Black, Beige/Turtledove/Incense, Black Neo Flame, Black Nero, Black/Black, Black/Magnet, Black/Magnet/Gum, Black/Moonbeam, Black/Shadow Grey/White, \ldots, Night Tide Aspen, Night Tide/Aspen, Nimbus Cloud/Crimson, Nimbus Cloud/Nightwatch Green, Nimbus Cloud/White, Nimbuscloud/Crimson, \ldots, White/Moonrock/Olivine, Workwear/Jade/Incense \emph{(57 values total)}
& correct
& All color values are distinct multi-color combinations with different descriptive elements. While `Night Tide Aspen' and `Night Tide/Aspen' appear similar, they have spacing differences that could represent different color arrangements. Similarly, `Nimbus Cloud/Crimson' and `Nimbuscloud/Crimson' have spacing differences, but these are not clear formatting duplicates as they could represent intentional product distinctions.
& egregious error
& After analyzing all 57 color values, I found one duplicate pair: `Nimbus Cloud/Crimson' and `Nimbuscloud/Crimson'. These differ only in spacing (space vs no space between `Nimbus' and `Cloud'), making them identical except for formatting. All other values represent genuinely different color combinations with distinct descriptive elements. Since at least one duplicate pair is present, this is classified as a duplicate. \\
\bottomrule
\end{tabularx}
\caption{Qualitative duplicates examples on Qwen3 8B where RBD labels the variation family correct but PTC detects the duplicate. In both cases the duplicate differs only by casing/spelling (\texttt{Dark Gray}/\texttt{Dark Grey}) or spacing (\texttt{Nimbus Cloud}/\texttt{Nimbuscloud}); RBD treats these as intentional variants, while PTC's explicit task conditioning flags them as duplicates.}
\label{tab:duplicates_qualitative}
\end{table*}

\begin{table}[t]
\centering
\scriptsize
\begin{tabular}{lll}
\toprule
\textbf{Model} & \textbf{PTC vs.\ method} & \textbf{p value} \\
\midrule
\multirow{2}{*}{Qwen3 8B}          & Non FT & $<0.0002$ \\
                & RBD    & $<0.0002$ \\
\midrule
\multirow{2}{*}{Mistral Nemo 12B}  & Non FT & $<0.0002$ \\
                                   & RBD    & $0.0024$ \\
\midrule
\multirow{2}{*}{Qwen3 14B}         & Non FT & $<0.0002$ \\
                                   & RBD    & $0.036$  \\
\midrule
\multirow{2}{*}{Mistral Small 24B} & Non FT & $<0.0002$ \\
                                   & RBD    & $0.0686$ \\
\midrule
\multirow{2}{*}{Qwen3 32B}         & Non FT & $<0.0002$ \\
                                   & RBD    & $0.0134$ \\
\bottomrule
\end{tabular}
\caption{Statistical significance of PTC's performance gains over the non-finetuned baseline and RBD, measured via a paired subsampled bootstrap test (5{,}000 class-balanced 80\% draws without replacement, plus-one correction) on the human-annotated test set. PTC is significant ($p < 0.05$) over RBD on four of five models.}
\label{tab:statistical_significance}
\end{table}

Table~\ref{tab:duplicates_qualitative} reveals that the gap between RBD and PTC on duplicate detection is not one of \emph{perception} but of \emph{execution}. In both examples, RBD's reasoning explicitly surfaces the offending pair. It names \texttt{Dark Gray}/\texttt{Dark Grey} and \texttt{Nimbus Cloud/Crimson}/\texttt{Nimbuscloud/Crimson} as candidates, so the relevant variants are clearly within the model's attention. Yet RBD then \emph{rationalizes them away}, recasting a spelling difference as ``different standardized spellings\ldots commonly accepted as distinct color variations'' and a spacing difference as something that ``could represent intentional product distinctions.'' Having identified the duplicate, RBD talks itself out of acting on the task. The error is therefore a failure to apply the task criterion, not a failure to notice the evidence.

PTC encounters the same lists and flags the same pairs, but converts the observation into the correct label. We hypothesize this stems from the task token placed immediately before the input: by re-anchoring the duplicate-detection objective at the point where the model transitions from instructions to data, PTC keeps the task criterion active while the candidate values are being read, rather than letting it be diluted across the preceding rule text. The effect is most visible in the 57-value \texttt{SHOES} list, where the lone \texttt{Nimbuscloud/Crimson} collision is buried among genuinely distinct multi-color combinations: PTC isolates the single offending pair and commits to the duplicate label, while RBD defers to the non-duplicate classification whenever any doubt remains. This pattern, repeated across the 29 duplicate cases that PTC recovers over RBD on Qwen3 8B, is consistent with our broader hypothesis that mid-prompt task reinforcement mitigates the attenuation of the initial task signal in long contexts.

\section{Statistical significance}
\label{sec:statistical_significance}

Table~\ref{tab:statistical_significance} reports the p-values from the paired subsampled bootstrap test described in \S\ref{sec:results}, comparing PTC against both the non-finetuned baseline (Non FT) and rationale-based distillation (RBD) on the human-annotated test set. PTC's improvement over the non-finetuned baseline is highly significant ($p < 0.0002$) across all five models. Against the stronger RBD baseline, PTC's gains are statistically significant ($p < 0.05$) on four of five models. The exception is Mistral Small 24B ($p = 0.0686$), where the improvement is directionally consistent but does not reach significance.

\end{document}